\documentclass[sigconf]{acmart} 

\AtBeginDocument{%
  \providecommand\BibTeX{{%
    \normalfont B\kern-0.5em{\scshape i\kern-0.25em b}\kern-0.8em\TeX}}}

\usepackage{soul}
\usepackage{color}
\usepackage{wallpaper}
\usepackage{graphicx}
\usepackage{cleveref}
\usepackage{numprint}
\usepackage{graphicx}
\usepackage{subcaption}
\usepackage{mwe}
\definecolor{orangeX}{rgb}{1,.5,0}
\definecolor{blueX}{rgb}{.2, .59, .88}
\definecolor{purpleX}{rgb}{.294118, 0, .509804}
\definecolor{greenX}{rgb}{.421, .578, .241}
\definecolor{bole}{rgb}{0.47, 0.27, 0.23}
\definecolor{mypink3}{cmyk}{0, 0.7808, 0.4429, 0.1412}
\definecolor{mygray}{gray}{0.6}
\usepackage{courier}
\newcommand{\name}{\textsc{MuSe-Toolbox}\,}

\newcommand{\musesec}{\textsc{MuSe-CaR}\,}

\newcommand{\ewe}{\textsc{EWE\,}}
\newcommand{\ewelong}{\emph{Estimator Weighted Evaluator\,}}
\newcommand{\dba}{\textsc{DBA\,}}
\newcommand{\dbalong}{\emph{DTW Barycenter Averaging\,}}
\newcommand{\gctw}{\textsc{GCTW\,}}
\newcommand{\gctwlong}{\emph{Generic-Canonical Time Warping\,}}
\newcommand{\awe}{\textsc{RAAW\,}}
\newcommand{\awelong}{\emph{Rater Aligned Annotation Weighting\,}}

\newcommand{\kmeans}{k-means\,}
\newcommand{\cmeans}{fuzzy c-means\,}
\newcommand{\gmm}{Gaussian Mixture Model\,}
\newcommand{\agglo}{agglomerative clustering\,}

\newcommand{\dbscan}{Density-Based Spatial Clustering of Applications with Noise \,}

\newcommand{\vgg}{\textsc{VGGish\,}}

\newcommand{\bert}{\textsc{BERT\,}}

\usepackage{balance} 

\newcommand{\eg}{e.\,g.,\,}
\newcommand{\ie}{i.\,e.,\,}

\newcommand{\cf}{{cf.\,}}

\copyrightyear{2021}
\acmYear{2021}
\setcopyright{acmlicensed}\acmConference[MuSe '21]{Proceedings of the 2nd Multimodal Sentiment Analysis Challenge}{October 24, 2021}{Virtual Event, China}
\acmBooktitle{Proceedings of the 2nd Multimodal Sentiment Analysis Challenge (MuSe '21), October 24, 2021, Virtual Event, China}
\acmPrice{15.00}
\acmDOI{10.1145/3475957.3484451}
\acmISBN{978-1-4503-8678-4/21/10}

\begin{document}
\fancyhead{}
\title{\textsc{MuSe-Toolbox}: The Multimodal Sentiment Analysis Continuous Annotation Fusion and Discrete Class Transformation Toolbox}


\author{Lukas Stappen}
\affiliation{%
  \institution{University of Augsburg}
  \city{Augsburg, Germany}}

\author{Lea Schumann}
\affiliation{%
  \institution{University of Augsburg}
  \city{Augsburg, Germany}}

\author{Benjamin Sertolli}
\affiliation{%
  \institution{University of Augsburg}
  \city{Augsburg, Germany}}
 
\author{Alice Baird}
\affiliation{%
  \institution{University of Augsburg}
  \city{Augsburg, Germany}}

\author{Benjamin Weigell}
\affiliation{%
  \institution{University of Augsburg}
  \city{Augsburg, Germany}}

  
\author{Erik Cambria}
\affiliation{%
  \institution{Nanyang Technological University}
  \city{Singapore}}  
  
\author{Bj\"orn W. Schuller}
\affiliation{%
  \institution{Imperial College London}
  \city{London, United Kingdom}}

\renewcommand{\shortauthors}{Stappen, et al.}

\begin{abstract}
We introduce the \name -- a Python-based open-source toolkit for creating a variety of continuous and discrete emotion gold standards. In a single framework, we unify a wide range of fusion methods and propose the novel \awelong (\awe), which aligns the annotations in a translation-invariant way before weighting and fusing them based on the inter-rater agreements between the annotations. Furthermore, discrete categories tend to be easier for humans to interpret than continuous signals. 
With this in mind, the \name provides the functionality to run exhaustive searches for meaningful class clusters in the continuous gold standards. To our knowledge, this is the first toolkit that provides a wide selection of state-of-the-art emotional gold standard methods and their transformation to discrete classes. 
Experimental results indicate that \name can provide promising and novel class formations which can be better predicted than hard-coded classes boundaries with minimal human intervention. The implementation\footnote{\url{https://github.com/lstappen/MuSe-Toolbox}} is out-of-the-box available with all dependencies using a Docker container\footnote{\texttt{docker pull musetoolbox/musetoolbox}}.

\end{abstract}

\ccsdesc[500]{Information systems~Multimedia and multimodal retrieval}
\ccsdesc[500]{Computing methodologies~Artificial intelligence}

\keywords{Affective Computing; Annotation; Gold-Standard; Smoothing; Emotion classes; Emotion Recognition; Multimodal Sentiment Analysis}

\maketitle

\vspace{-0.3cm}
\section{Introduction}
The accelerating pace of digitisation is driving digital interaction into all areas of our daily life, and from the resulting mass of data, a substantial portion can be quantified into human behavioural signals. 
Learning to recognise emotional cues in interactions \eg taking place via video, is the purpose of the growing field of Emotion AI. In this process, various modalities, such as body language, voice, text, and facial expression, are examined for patterns that help map the cues to specific emotions. The reference data necessary to learn the mapping is annotated by humans, often as category labels (\eg{} happy, sad, surprised, etc.) and continuous annotations. For continuous mapping, behavioural and cognitive scientists assume that the human brain is not divided into hard-wired regions and better represented by dominant primitives (dimensions) whose complex interaction results in a specific emotion (\eg the dimensional axes of arousal and valence) \cite{russell1980_Circumplex}.

The growing demand for emotion technology in various domains led to an increased interest in the annotation of such data. However, the annotation process itself is not trivial to execute, and obtaining meaningful reference data to develop models for automatic pattern recognition is a challenge. One such challenge is the dependency on humans raters. When rating the perceived data (\eg videos), time-delays in the reaction \cite{nicolaou2014dynamic}, as well as systematic disagreement due to personal bias and other task-related reasons
are well known~\cite{booth2018novel, atcheson2018demonstrating}. 
To counteract, it is common practice to involve multiple humans in the annotation of the same source and fuse these perceptions. Since emotions are inherently subjective, these fused signals are coined as gold-standard.
To date, none of the proposed fusion methods has become a de-facto standard. One reason for this may be that a convenient comparison of the fusion outcomes is hardly feasible. The implementation of the methods is often distributed over many different source bases, coded in different programming languages and frameworks, or is not publicly available at all. An issue is also the dependency on outdated software (package) dependencies.

Another unresolved problem is the transformation of continuous emotion signals into more general class labels that are easier for humans to interpret. In an empirical approach, Hoffmann et al.~\cite{Hoffmann_2012} mapped discrete emotions into the dimensional emotion space \cite{russell1980_Circumplex}. Similarly, Laurier et al.~\cite{Laurier_2009} aimed to cluster emotion tags to find clusters corresponding to the four quadrants of the arousal-valence dimensions.
A tool that supports this transformation process by the automatic creation of meaningful classes has not yet been presented in the literature.

With this contribution, we want to tackle both of these issues by proposing an easy-to-use, well-documented toolbox. The input data can be any continuous annotation recorded by an annotation software (\eg{} a human-controlled joystick or mouse) or directly from a (physiological) device (\eg{} smartwatches). Additionally, the annotations can be easily standardised, smoothed and fused by the most common gold-standard creation techniques,  such as \ewelong (EWE), \dbalong (DBA), and \gctwlong (GCTW). This elegantly makes a comparison of the multiple available fusion methods easily possible, leaving broad flexibility for database creators while allowing reproducibility and exchange over the set of parameters used. To this end, we propose a novel gold-standard method \awelong (\awe) to the set of fusion tools, which we derived from methods introduced here and which is inspired by the results we obtained during the work on the toolbox and the limitations of the provided fusion methods. Furthermore, we propose a simple way to extract time-series features from these signals, which may aid the creation of emotional classes from emotion dimensions. The toolbox can be started directly from a Docker container without installing dependencies, and an open-source Github repository is available to the community for further development.

Note, the core focus of this work is emotional annotations. However, all kinds of time-series data are omnipresent in our daily life. Changes in stocks, energy consumption, or weather are all recorded over time and, thus, have natural time-series properties. Predicting these values in time is often challenging and a simplification by fusing them (\eg energy consumption of several households) transforming sequences into summary classes by clustering (\eg days in a week) may be beneficial for any of the other applications as well. 

\vspace{-0.2cm}
\section{Methodology and System Overview}
In the following section, we first describe the methods that underpin the functionality of our toolbox and conclude by placing them in the context of the functionalities in \Cref{sec:box_overall}.

\vspace{-0.2cm}
\subsection{Smoothing of Annotations\label{sec:smoothnorm}} 
\begin{figure}[t!]
    \centering
    \includegraphics[width=.45\textwidth]{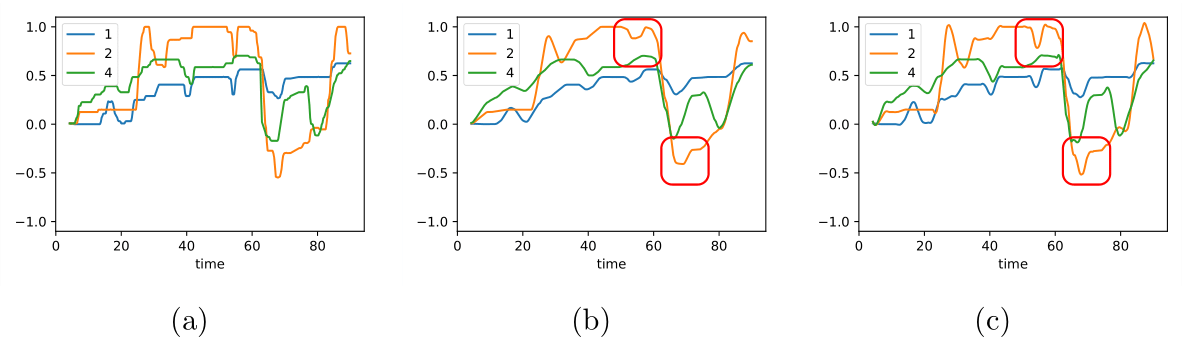}
    \caption{An example of valence annotation signals of three annotators. Figure (a) depicts the raw signals, while the other two figures show the filtered signals of a moving average filter (b), and a cubic Savitzky-Golay filter (c), respectively, with a filter frame size of 17 values (4.25\,s). Evidently, the moving average indicates a visibly stronger smoothing effect, when compared to the Savitzky-Golay filter, which preserves signal features more closely.} 
    \label{fig:annow_raw}
    \vspace{-0.1cm}
\end{figure}

As for all fine-grained time-series, short-term errors and distortions can occur in the annotation process. Smoothing digital filters are useful to mitigate these negative noise effects~\cite{thammasan2016investigation, wang2018towards}. One common signal processing approach for this is the \emph{Savitzky-Golay filter} (SavGol) which increases the precision of the data points using a low degree polynomial over a moving filter~\cite{savitzky1964smoothing}. In our context, this method has the advantage that it still preserves high-frequency characteristics ~\cite{thammasan2016investigation}. Also widely applied is the \emph{Moving Average Filter} (MAF). It employs a moving average of a given window to smooth the signal gently. The MAF applied with $4.25\,s$ filter frame (or 17 time steps) is illustrated in \Cref{fig:annow_raw}, alongside a SavGol example, and the raw annotations.


\vspace{-0.1cm}
\subsection{Gold Standard Fusion Methods}
\begin{figure}[t!]
    \centering 
    \includegraphics[ width=0.49\columnwidth]{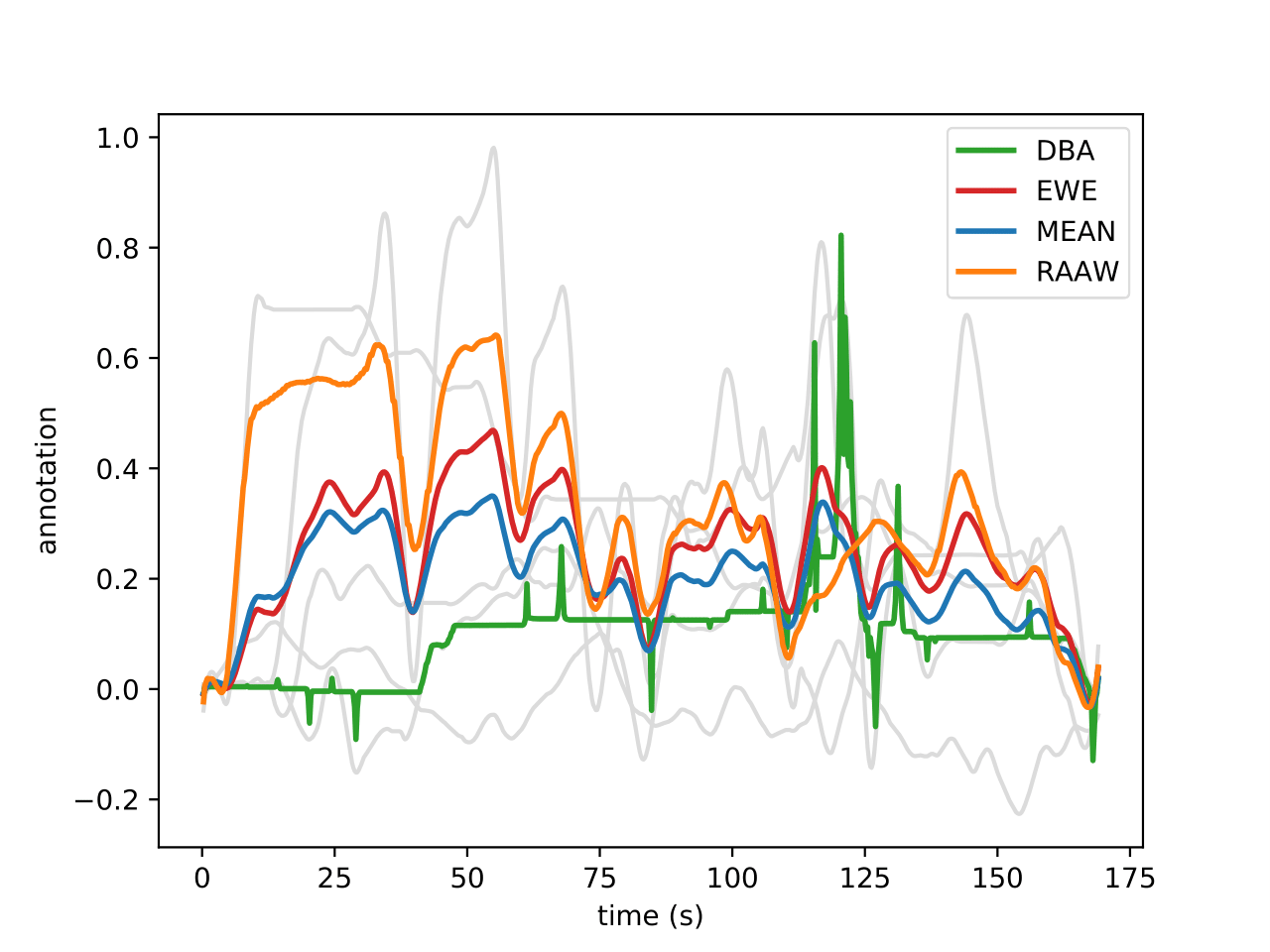}
    \includegraphics[width=0.49\columnwidth]{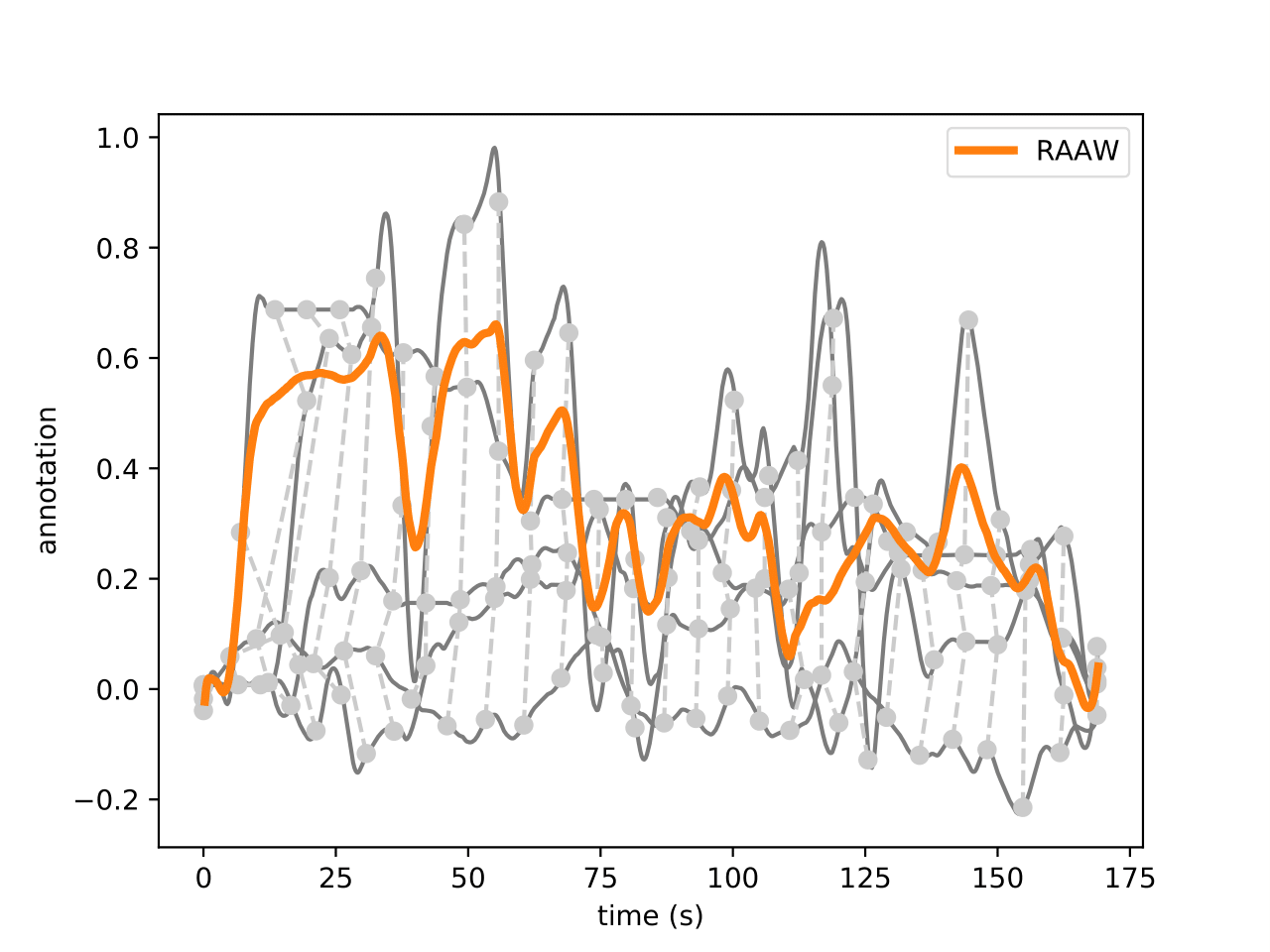}
    \caption{The left side shows all fusion methods in \name{} on a sample annotation (\musesec{} database, video id: 100, arousal). The right side is a detailed illustration of the \awelong{} (\awe) alignment, including the warping paths.}
    \label{fig:methods}
\end{figure}

A gold-standard method tries to establish a consensus from a group of individual ratings. Some methods are specifically developed for emotion annotations, \ie \ewe, and \awe, while others are derived from more generic principles of time-series aggregation, \ie \dba, \gctw.  A comparison of all methods is visualised in \Cref{fig:methods}.

\subsubsection{Estimator Weighted Evaluator (EWE)\label{sec:ewe}}
The \ewelong (\ewe) is based on the reliability evaluation of the raters~\cite{schuller2013intelligent}. It is essentially a weighted mean of all rater-dependent annotations, sometimes interpreted as the weighted mean of raters' similarity~\cite{grimm2005evaluation, hantke2016introducing}. To compute the weights, the cross-correlations of an annotation to the mean of all other annotations is calculated for each annotation. It can be formally expressed by
\begin{equation}
\hat{x}_{n}^{E W E}=\frac{1}{\sum_{k=1}^{K} r_{k}} \sum_{k=1}^{K} r_{k} \hat{x}_{n, k},
\end{equation}
where $r_k$ is the similarity of the k-th annotator to the other time-series. A typical method for calculating similarity between time-series is the Euclidean metric or Pearson coefficients. However, since both do not take sequence order, phase shift, and scaling variance into account, it was replaced by the concordance correlation coefficient (CCC) for similarity calculation: 
\begin{equation}
CCC(\hat{\theta}, \theta) = \frac{2 \times COV(\hat{\theta}, \theta)}{\sigma^2_{\hat{\theta}} + \sigma^2_\theta + (\mu_{\hat{\theta}} - \mu_\theta)^2} = \frac{2E[(\hat{\theta}-\mu_{\hat{\theta}})(\theta-\mu_{\theta})]}{\sigma^2_{\hat{\theta}} + \sigma^2_\theta + (\mu_{\hat{\theta}} - \mu_\theta)^2}.
\end{equation}
Here, $x$ is the time-series data, $\theta$ is a series of $n$ annotations,  and $\hat{\theta}$ the reference annotation. This method is broadly applied across different tasks in affective computing~\cite{ringeval2013introducing, ringeval2017avec,kossaifi2019sewa,stappen2020muse}. 

\subsubsection{DTW Barycentre Averaging (DBA)\label{sec:dba}}

Averaging in Dynamic Time Warping (DTW) spaces is widely adopted for similarity-based temporal alignment in the field of machine learning. Similar to the Euclidean metric and CCC, DTW implements a distance metric, adding elastic properties that compute the best global alignment based on a one-to-many mapping of points in two time-series. 
The \dbalong (\dba) method available in our framework is based on an algorithm originally developed for general time-series barycentre computation to compute the optimal average sequence. A barycentre is a time-series $b$ based on the computation of continuous representative propensities from multiple time series points $x$. In this particular version, these tendencies are determined by a sub-gradient, majorize-minimize algorithm of $d$ \cite{schultz2018nonsmooth} with the advantage of fusing of time-series of varied length. DTW can be expressed as:
\begin{equation}
DTW = \min \sum_{i} d\left(b, x_{i}\right)^{2}.
\end{equation}


\subsubsection{Generic-Canonical Time Warping (GCTW)\label{sec:gctw}}
Another extension of DTW is Canonical Time Warping (CTW)~\cite{zhou2009canonical}, which in addition to DTW integrates Canonical Correlation Analysis~\cite{anderson1958introduction}, a method for extracting shared features from two multi-variate data points. CTW was originally developed with the goal of aligning human motion and multimodal time series more precisely in time~\cite{zhou2009canonical}. The combination with these two approaches allows a more flexible way of time-warping by adding monotonic functions that can better handle local spatial deformations of the time series. The same authors~\cite{zhou2015generalized} further extended this approach to \gctwlong (\gctw), which enables a computationally efficient fusion of multiple sequences by reducing the quadratic to linear complexity. Furthermore, the identified features with high correlation are emphasised by weighting. 


\subsubsection{\awelong (\awe)\label{sec:awe}}

In the context of emotions, we propose a novel method \awelong (\awe) for the fusion of dimensional annotations for gold-standard creation. \awe capitalises on the merits of the underlying alignment technique DTW and the inherent nature of the EWE method. More specifically, DTW is used to align the varying and changing response times of individual annotators over time (\cf \Cref{fig:methods}).
This alignment between the fused signal was previously made brute-force by shifting the global or individual annotation by a few seconds and measure the resulting performance.
The optimal number of emotion annotators is estimated to be at least three depending on their experience and the difficulty of the task~\cite{honig2010many}. To perform an alignment in a resource-efficient manner --- even for many annotations --- we utilise the DTW variant GCTW~\cite{zhou2015generalized}. Subsequently, the similarity is calculated using the CCC for the individual aligned signals to accommodate the inter-rater agreement (subjectivity). The signals weighted according to this can be completely disregarded when negatively correlated before they are finally merged using EWE~\cite{grimm2005evaluation}. 

\vspace{-0.1cm}
\subsection{Emotional Signal Features\label{sec:signalfeatures}}
Emotion annotations can be seen as a quasi-continuous signal with a high sampling rate \cite{kossaifi2019sewa, stappen2020summary, stappen2020muse}. Extracting features from audio-visual and psychological signals is fairly common in intelligent computational analysis~\cite{schuller2013intelligent, schuller2020interspeech}. In the context of this work, we extract (time-series) features from an emotional signal segment to summarise the time period in a meaningful way. The resulting representation summarising the segment over time is a vector of the size of the selected features. Starting with the most interpretable features, common statistical measures are extracted~\cite{Sagha17-PTP}, such as the standard deviation ($std$), mean, median and a range of quantiles ($q_x$). However, these features do not reflect the characteristics of changes over time.

For this reason, the toolkit further offers to extract more complex time-series features namely: relative energy (\textit{relEnergy})~\cite{christ2018time}, mean absolute change (\textit{MACh}), mean change (\textit{MCh}), mean central approximation of the second derivatives (\textit{MSDC}), relative crossings of a point $m$ (\textit{CrM})~\cite{christ2018time}, relative number of peaks (\textit{relPeaks})~\cite{christ2018time,palshikar2009simple}, skewness~\cite{doane2011measuring,ekman1992argument}, kurtosis~\cite{westfall2014kurtosis}, relative longest strike above the mean (\textit{relLSAMe}), relative longest strike below the mean (\textit{relLSBMe}), relative count below mean (\textit{relCBMe}), relative sum of changes (\textit{relSOC}), first and last location of the minimum and maximum (\textit{FLMi}, \textit{LLMi}, \textit{FLMa}, \textit{LLMa}), and percentage of reoccurring data points ($PreDa$).
Note that features labelled as ``relative'' are normalised by the length of a segment, in order to limit the influence of varying segment lengths on the unsupervised clustering.

\vspace{-0.1cm}
\subsection{Dimension Reduction\label{sec:dimreduc}}
Large dimensional feature sets often lead to unintended side effects, such as the curse of dimensionality~\cite{trunk1979problem}. However, by reducing or selecting certain dimensions of the available features, these effects can be counteracted. 
Principal component analysis (PCA) is a well-known dimension reduction method that transforms features into principal components~\cite{Zaki:31:PCA}. These components are generated by projecting the original features into a new orthogonal coordinate system. This enables the reduction of the dimensions while preserving most of the data variation.
Another method for dimensionality reduction is Self-organising Maps (SOM), a type of unsupervised, shallow neural network that transforms a high-dimensional input space into a low-dimensional output space~\cite{Kohonen:1990}. Each output neuron competes with the other neurons to represent a particular input pattern, which makes it possible to obtain a comprehended representation of the most relationships in the dataset. SOM can also be used as a clustering or visualization tool, as they are considered to have low susceptibility to outliers and noise~\cite{Vesanto:7}. 
\vspace{-0.1cm}
\subsection{Clustering\label{sec:cluster}}

\subsubsection{K-means and fuzzy c-means clustering\label{sec:kmeans}}
A common way to differentiate \kmeans from \cmeans algorithms is how a datapoint belongs to the resulting outcome, which can either be an assignment to exactly one cluster (crisp), or to multiple ones with a certain probability (fuzzy). The most popular fuzzy clustering method is the \cmeans algorithm~\cite{Bezdek:1984}, based on the \kmeans algorithm~\cite{Hartigan:1979}. To this end, a fixed number of clusters is defined. The cluster centres are initially set randomly, and the Euclidean distances from them to the data points are calculated. These are assigned to the clusters so that there is a minimal variance increase. By step-wise optimisation (similar to an expectation maximisation (EM) algorithm) of the centres and assignments, the algorithm converges after a few iterations. For the fuzzy version, the degree of overlap between clusters can be specified using the fuzzifier $m$ parameter. 

\vspace{-0.1cm}
\subsubsection{Gaussian mixture model \label{sec:gmm}}
Similar to c-means, a \gmm (GMM) introduces fuzziness into the clustering process and allows the weak assignment of a single datapoint to several clusters simultaneously. For this purpose, a probabilistic model is generated that attempts to describe all data by Gaussian distributions with different parameters. The optimisation process to find a suitable covariance structure of the data as well as the centres of the latent Gaussian distributions uses the EM algorithm as in \kmeans.

\vspace{-0.1cm}
\subsubsection{Agglomerative clustering\label{sec:agllo}}
Besides the \kmeans, two other types of crisp clustering are common: agglomerative~\cite{Kaushik:26} and density clustering. Agglomerative is a hierarchical clustering technique in which each datapoint starts as its own cluster and is successively merged with the closest datapoint (\ie cluster) into higher-level clusters. As soon as the distance between two clusters is maximised or the minimum number of clusters is reached, the clustering process is terminated. 
\vspace{-0.1cm}
\subsubsection{Density-Based Spatial Clustering of Applications with Noise (DBSCAN)\label{sec:dense}}
Density-clustering algorithms such as \dbscan (DBSCAN)
have became more popular over the last years~\cite{Campello:2013}. The main difference to other methods is that it also uses the local density of points instead of relying only on distance measures~\cite{Ester:1996}. DBSCAN provides an answer to two common problems in clustering: a) the number of clusters does not have to be specified in advance and b) it automatically detects outliers which are then excluded from the clustering~\cite{Sharma:25,Kaushik:26}. With other methods, these outliers have to be removed manually after a manual check, otherwise, there is a risk that the clusters would get distorted. The reason for this is that each point must contain at least a minimum number of points in a given radius, called min\_samples parameter in the $\epsilon$-neighborhood. However, this simultaneously causes a firm reliance on the defined parameters.
\vspace{-0.1cm}
\subsubsection{ Measures\label{sec:measures}}
Clusters are usually evaluated using internal metrics and external assessment. The internal metrics focus on how similar the data points of a cluster are (compactness), and how far the clusters differ from each other (separation) \cite{Liu:6}. 
The Calinski-Harabasz Index (CHI) calculates the weighted average of the sums of squares within and between clusters. 
Also distance-based is the Silhouette Coefficient (SiC), but it is bounded within an interval of -1 to 1 (1 corresponds to an optimal cluster), allowing for easier comparability between runs and procedures~\cite{Zaki:31:S}. The Davies-Bouldin Index (DBI) is based on similarity measures and decreases with increasing cluster separability~\cite{Zaki:31:DB}. Specifically for \cmeans, the Fuzzy Partition Coefficient (FPC) can be employed, and measures the separability of \cmeans using Dunn's partition coefficients \cite{Dunn}. Finally, we use the S\_Dbw-Index, which is based on intra-cluster variance to measure compactness, where the average density in the area between clusters and the density of clusters is calculated (smaller is better).

\vspace{-0.1cm}
\subsection{MuSeFuseBox System Overview\label{sec:box_overall}} 
\vspace{-0.1cm}
\begin{figure}[ht!]
	\centering
	\includegraphics[width=\columnwidth]{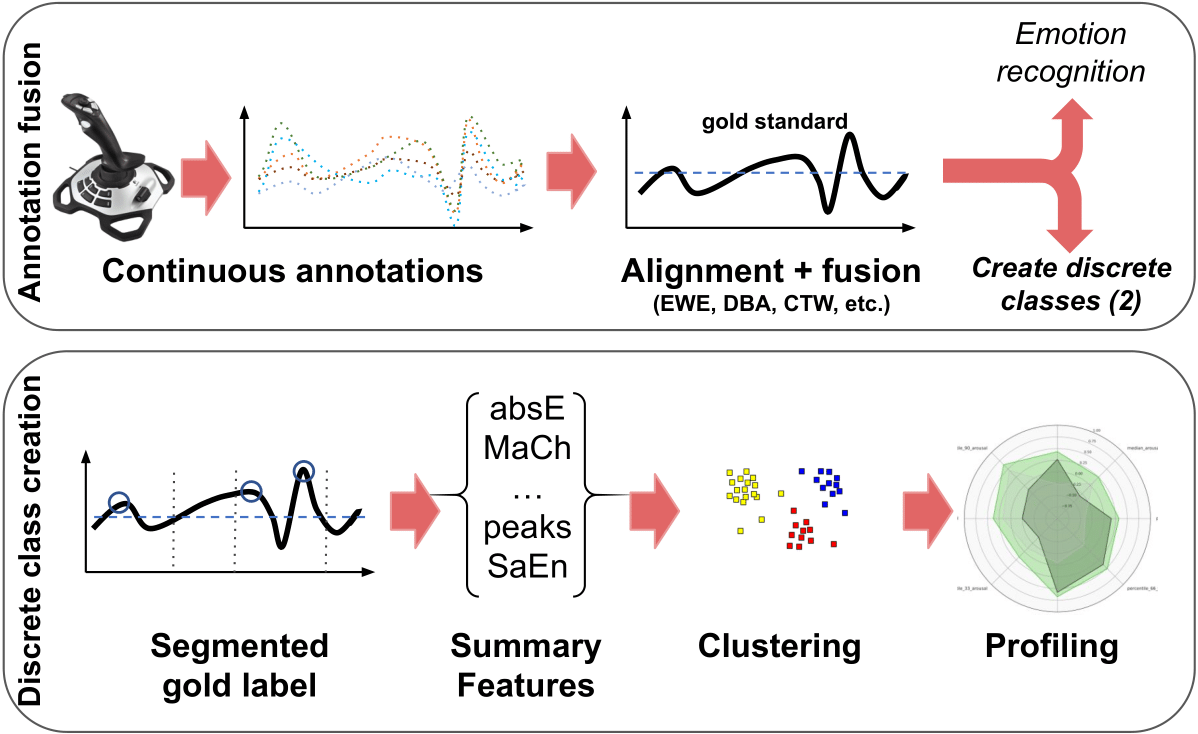}
	\caption{System overview of \name}
	\label{fig:overview}
\end{figure}

The introduced methodology is integrated into the \name as depicted in \Cref{fig:overview}. The upper part shows the annotation fusion process. Given the input of multiple annotations, these can first be smoothed and/or normalised (\cf \Cref{sec:smoothnorm}), which has shown benefits in previous works~\cite{martinez2020msp, ringeval2017avec}. The normalisation is either applied on video- or annotator-level. Next, the pre-processed annotations are fused using either \dba, \ewe, \gctw, or \awe (\cf \Cref{sec:smoothnorm}). The lower part represents the creation process of discrete classes from a given signal. All, or a selection of the introduced features from \Cref{sec:signalfeatures} are extracted from segments of the fused annotation signal. These summary features are either clustered directly by one of the methods described in \Cref{sec:cluster} or first reduced in dimensionality (\cf \Cref{sec:dimreduc}) and subsequently clustered. There is an option to either cluster on all data or on the training partition only. For the latter option, the classes of the development and test partitions are predicted based on the resulting clusters from the training set. For internal evaluation, the measures described in~\Cref{sec:measures} are calculated. Since the generated clusters are intended to be used as classification targets, an exclusion of clustering proposals based on a rule-of-thumb can be activated to avoid strong class imbalances. This excludes cluster proposals where one or more clusters are smaller than a factor of the by chance level. For example, the prediction of four classes has a by chance level of 25\,\%. If the factor is set to 0.5, then, the smallest proposed cluster has to cover at least 12.5\,\% of the data. Finally, the profiling provides all the information necessary to enable an additional external evaluation by a human. For profiling, we provide a) standard features (mean, standard deviation, etc.), b) visualisations, such as radar charts of the top distinctive features and scatter plots, and c) correlation of the features within a cluster. Based on these, the resulting clusters can be interpreted, and a name can be given.

\vspace{-0.1cm}
\subsection{Implementation Details}
The \name is implemented in Python and relies on several packages, most notably numpy, pandas, scikit-learn, oct2py, and scipy. It can be used as a command line tool (over 50 different settings and configurations are available) or from the Python API. The implementation of DBA is adapted from \cite{Petitjean2011-DBA,Petitjean2014-ICDM-2,Forestier2017-ICDM}\footnote{\url{https://github.com/fpetitjean/DBA},
GNU General Public License} and DTW components are adapted from the Matlab implementation\footnote{\url{https://github.com/zhfe99}, free for research use (no licence)} of \cite{zhou2015generalized}, which we transformed into code of the open-source programming language and environment for octave and access it for our calculations. The code is publicly available on GitHub under the GNU General Public license\footnote{\url{https://github.com/lstappen/MuSe-Toolbox}}.

\section{Experiments} 
To demonstrate the capabilities of the \name, we run experiments on the produced gold standards. By doing so, we used them to train models for dimensional affect recognition. To this end, we utilise the \musesec database \cite{stappen2021multimodal}, used in the 2020 and 2021 Multimodal Sentiment Analysis real-life media Emotion Challenges (MuSe) \cite{stappen2020muse,stappen2021muse}, and several other works ~\cite{stappen2021sentiment,stappen2021estimation,stappen2021unsupervised,sun2020multi,fu2020aaec,li2020multi}. 

\vspace{-0.1cm}
\subsection{Continuous Emotion Fusion}

In this section, we present the results of several experiments based on outputs from our toolkit to demonstrate its functionality. As explained in the previous sections, gold-standard methods lead to qualitatively different results, meaning that the quantitative results alone are only of limited value.

For our experiments, we build on the MuSe \cite{stappen2020muse,stappen2021muse}, a challenge-series co-located to the ACM Multimedia Conference, which aims to set benchmarks for the prediction of emotions and sentiment with deep learning methods in-the-wild. Since the experimental conditions are predefined and publicly available, this is an ideal test ground. The database utilised for the challenge is called \musesec, which provides 40 hours of YouTube review videos of human-emotion interactions. Each 250 ms of the video dataset is labelled by at least five annotators, which are used for the following experiments. For more information, we refer the interested reader to the challenge \cite{stappen2021muse} and database paper \cite{stappen2021multimodal}.

We use two of the provided feature sets, \vgg and \bert, from the challenge \cite{stappen2021muse} to predict arousal and valence. \vgg{}~\cite{hershey2017cnn}, is a 128 dimensional audio feature set pre-trained on an audio dataset including YouTube snippets (AudioSet) with the aid of deep learning methods. These audio samples were differentiated into more than 600 different classes. \bert~\cite{devlin2019bert}  embeds words in vectors by using transformer networks. Its deep learning architecture is upfront trained on several datasets and training tasks. The embeddings used here is the sum of the last four output layers, which consists of a total of 768 dimensions. Both embeddings were extracted at the same sample rate as the labels. Furthermore, the LSTM-RNN baseline model made available by the organisers is utilised and re-trained for 100 epochs with batch size 1024 and learning rate $lr=0.005$ on the new targets. Further, we run a parameter optimisation for the hidden state dimensionality $h = \{32, 64\}$ for arousal and $h = \{64, 128\}$ to predict valence, as this selection has previously worked well for the \musesec data, as shown in the 2021 MuSe Challenge baseline publication \cite{stappen2021muse}. As the challenges use the CCC as the competition measure, we use the CCC for evaluation as well as the loss function. 

\subsubsection{Smoothing}

\begin{table}[t!]
\caption{Results comparing with and without pre-smoothing using a savgol filter with a size of 5 on all annotation fusion techniques.}
\resizebox{\linewidth}{!}{
    \begin{tabular}{@{}lrr|rr||rr|rr@{}}
    \toprule
    & \multicolumn{4}{c||}{\textbf{Arousal}}  &  \multicolumn{4}{c}{\textbf{Valence}}\\ 
    \midrule
               & \multicolumn{2}{c|}{--} & \multicolumn{2}{c}{smooth} & \multicolumn{2}{c}{--} & \multicolumn{2}{c}{smooth}  \\ 
               & \multicolumn{1}{l}{Devel.} & \multicolumn{1}{l|}{Test} & \multicolumn{1}{l}{Devel.} & \multicolumn{1}{l||}{Test} & \multicolumn{1}{l}{Devel.} & \multicolumn{1}{l|}{Test} & \multicolumn{1}{l}{Devel.} & \multicolumn{1}{l}{Test} \\ 
               \midrule
    \textbf{DBA}         & .2634     & .2615     & .2368    & .2480    & .3580     & .4209     & .2583    & .3638 \\
    \textbf{GCTW}        & .4809     & .3481     & .4840    & .3502    & .4394     & .5594     & .4503    & .5848 \\
    \textbf{EWE}         & .4410        & .2513  & .4386    & .3210     & .4476    & .5614     & .4454    & .5703 \\
    \textbf{RAAW}         & .4266       & .2778   & .4225    & .3514     & .4589    & .5493     & .4482    & .5698 \\
    \hline
    $\varnothing$      & .4030        &  .2847   & .3955    & .3177      & .4260   & .5228    & .4006    & .5222 \\
    \bottomrule
    \end{tabular}
}
\label{tab:Smoothing}
\end{table}

The effect of smoothing can be seen in~\Cref{fig:annow_raw}, c) compared to the raw annotations and the filtered signal using the Savitzky-Golay filter. It is apparent that the moving average filter smooths the signal much more than the Savitzky-Golay filter, even to a point at which information from the signal is lost. Hence, we adjust the filter frame-size of the moving average filter to be a smaller value compared to the Savitzky-Golay filter.
Following the pre-processing and fusion, the fused signal can further be smoothed using convolutional smoothing. The kernel size of 15 has proven to yield high-quality gold standard annotations whilst reduced signal noise.
We further compare the performance of all fusion methods when applying the Savitzky-Golay filter for pre-smoothing in \Cref{tab:Smoothing}. In general, it is noticeable that the DBA results are considerably below the level of the other three models. When predicting arousal, the models tend to overfit, while underfitting can be observed for the prediction of valence. This was also found in \cite{stappen2020muse, li2020multi, sun2020multi} and is possibly due to the chosen data split, which is speaker-independent, hence leading to imbalances in the label distribution \cite{stappen2021multimodal}.
For arousal, the results without normalisation are slightly stronger on the development set. On the test set, the overfitting gap for \ewe and \awe decreases by at least by .07 CCC with the application of the pre-smoothing filter. For valence, the results without the pre-smoothing filter are also slightly better on the development set, with the exception of \gctw. Pre-smoothing, however, produces atypically low results for DBA, which may indicate the sensitivity of the fusion method. With the other methods, the test result improved moderately.

\subsubsection{Normalisation}
\begin{table}[t!]
\caption{Results comparing different standardisation techniques (no pre-smoothing) on all annotation fusion techniques.}
\resizebox{\linewidth}{!}{
    \begin{tabular}{@{}lrr|rr|rr||rr|rr|rr@{}}
    \toprule
    & \multicolumn{6}{c||}{\textbf{Arousal}}  &  \multicolumn{6}{c}{\textbf{Valence}}\\ 
    \midrule
               & \multicolumn{2}{c|}{--} & \multicolumn{2}{c|}{per video} & \multicolumn{2}{c||}{per annotator} & \multicolumn{2}{c|}{--} & \multicolumn{2}{c|}{per video}  & \multicolumn{2}{c}{per annotator} \\ 
               & \multicolumn{1}{l}{Devel.} & \multicolumn{1}{l|}{Test} & \multicolumn{1}{l}{Devel.} & \multicolumn{1}{l|}{Test} & \multicolumn{1}{l}{Devel.} & \multicolumn{1}{l||}{Test}  & \multicolumn{1}{l}{Devel.} & \multicolumn{1}{l|}{Test} & \multicolumn{1}{l}{Devel.} & \multicolumn{1}{l|}{Test} &  \multicolumn{1}{l}{Devel.} & \multicolumn{1}{l}{Test} \\ 
               \midrule
    \textbf{DBA}         & .2811 & .1993 & .3616 & .2685 & .2634 & .2615 & .3072 & .2868 & .3580 & .4209 & .2800 & .3991 \\
    \textbf{GCTW}        & .4969 & .3558 & .5175 & .3207 & .4809 & .3481 & .4353 & .5345 & .4256 & .5170 & .4394 & .5594 \\
    \textbf{EWE}         & .4750 & .3563 & .4923 & .2746 & .4410 & .2513 & .4452 & .5551 & .4479 & .5193 & .4476 & .5614 \\
    \textbf{RAAW}         & .4546 & .2814 & .4898 & .3817 & .4266 & .2778 & .4411 & .5326 & .4430 & .5568 & .4589 & .5493 \\
    \hline
    $\varnothing$        & .4269 & .2982 & .4653 & .3114 & .4030 & .2847 & .4072 & .4773 & .4186 & .5035 & .4065 & .5173 \\
    \bottomrule
    \end{tabular}
}
\label{tab:Normalisation}
\end{table}

Across all methods, the maximum deviation on test of the average results is low at .02 CCC for arousal and .04 CCC for valence (\cf \Cref{tab:Normalisation}). On an individual level,  there are stronger differences, \eg the results for the fusion of arousal with \awe differ by more than .05 CCC on the development set and .1 CCC on the test set, with clear advantages for standardisation at the video level. This is the case for most gold standard procedures in predicting arousal (development set). The results for the prediction of valence are predominantly highest when standardised on the annotator level.

\begin{figure*}
    \centering
    \begin{subfigure}[b]{0.24\textwidth}
        \centering
        \includegraphics[trim={4cm 0 0 0},clip,width=\textwidth]{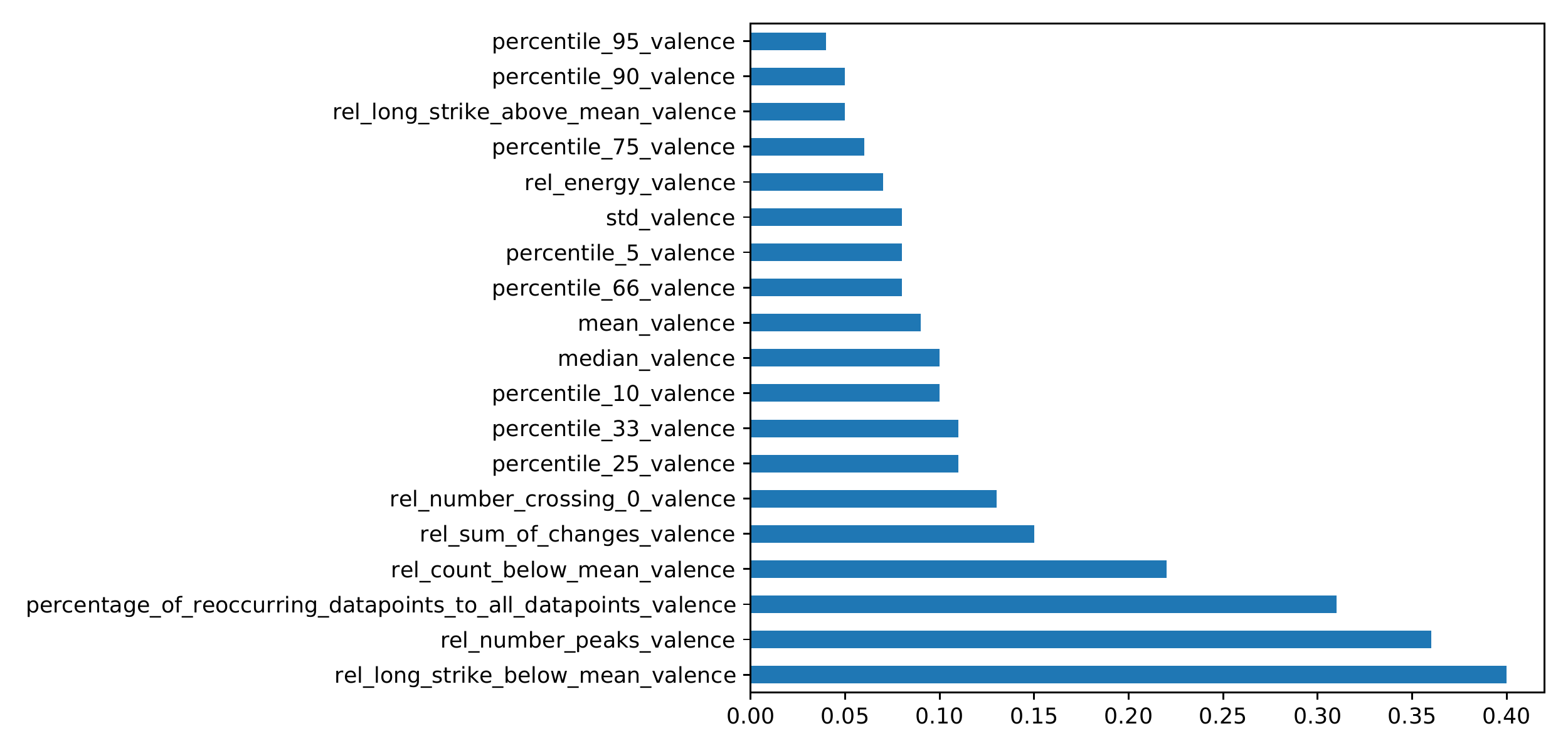}
        \caption[Network2]%
        {{\small Absolute correlation between the labels and all other features.}}    
        \label{fig:sub_corr}
    \end{subfigure}
    \hfill
    \begin{subfigure}[b]{0.24\textwidth}  
        \centering
        \includegraphics[trim={0 0 -1.2cm 0},clip,width=\textwidth]{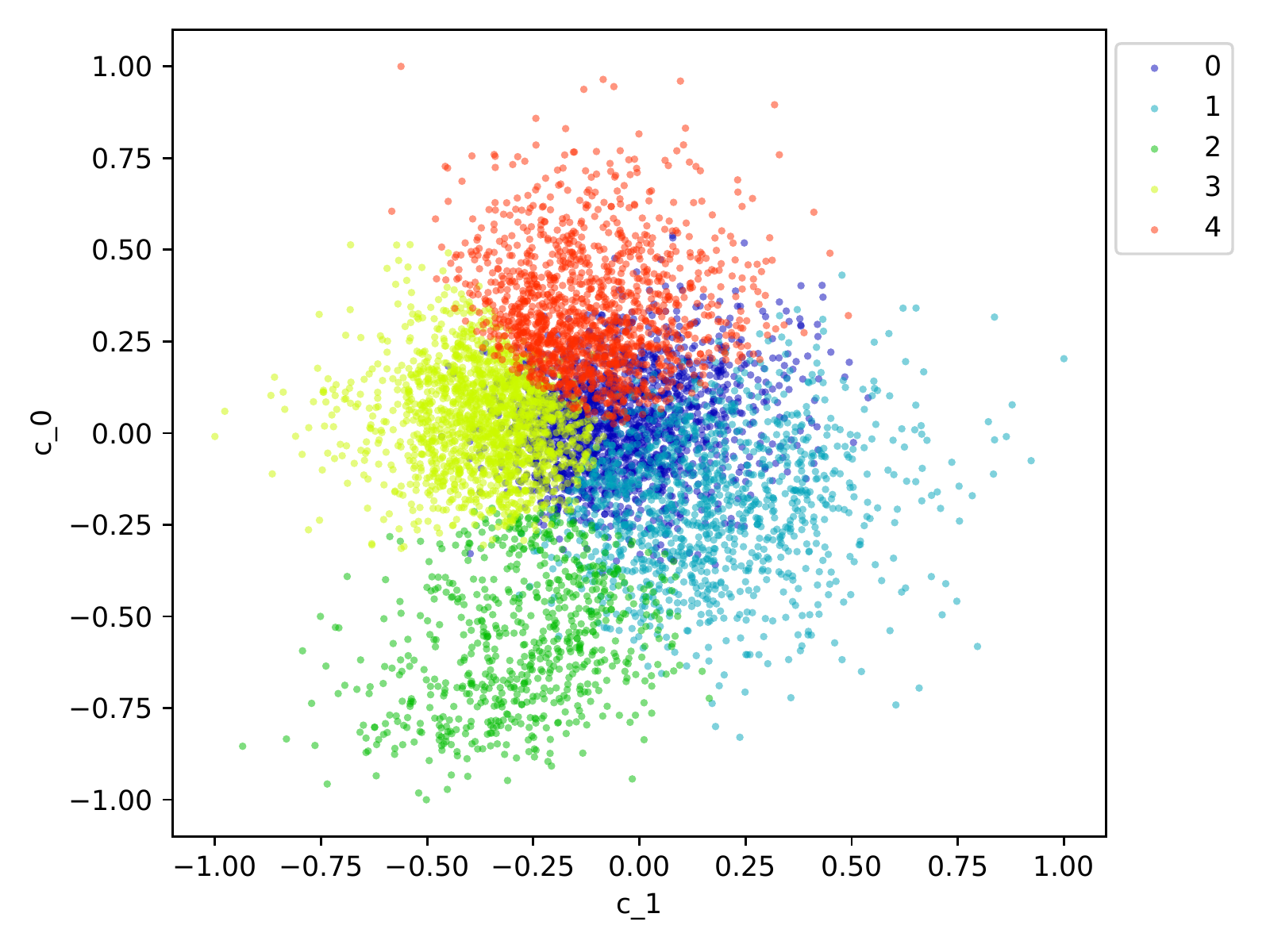}
        \caption[]%
        {{\small Cluster classes visualisation\\ after post-PCA.}}    
        \label{fig:sub_cloud}
    \end{subfigure}
    \begin{subfigure}[b]{0.24\textwidth}   
        \centering 
        \includegraphics[trim={0 0 0 0},clip,width=\textwidth]{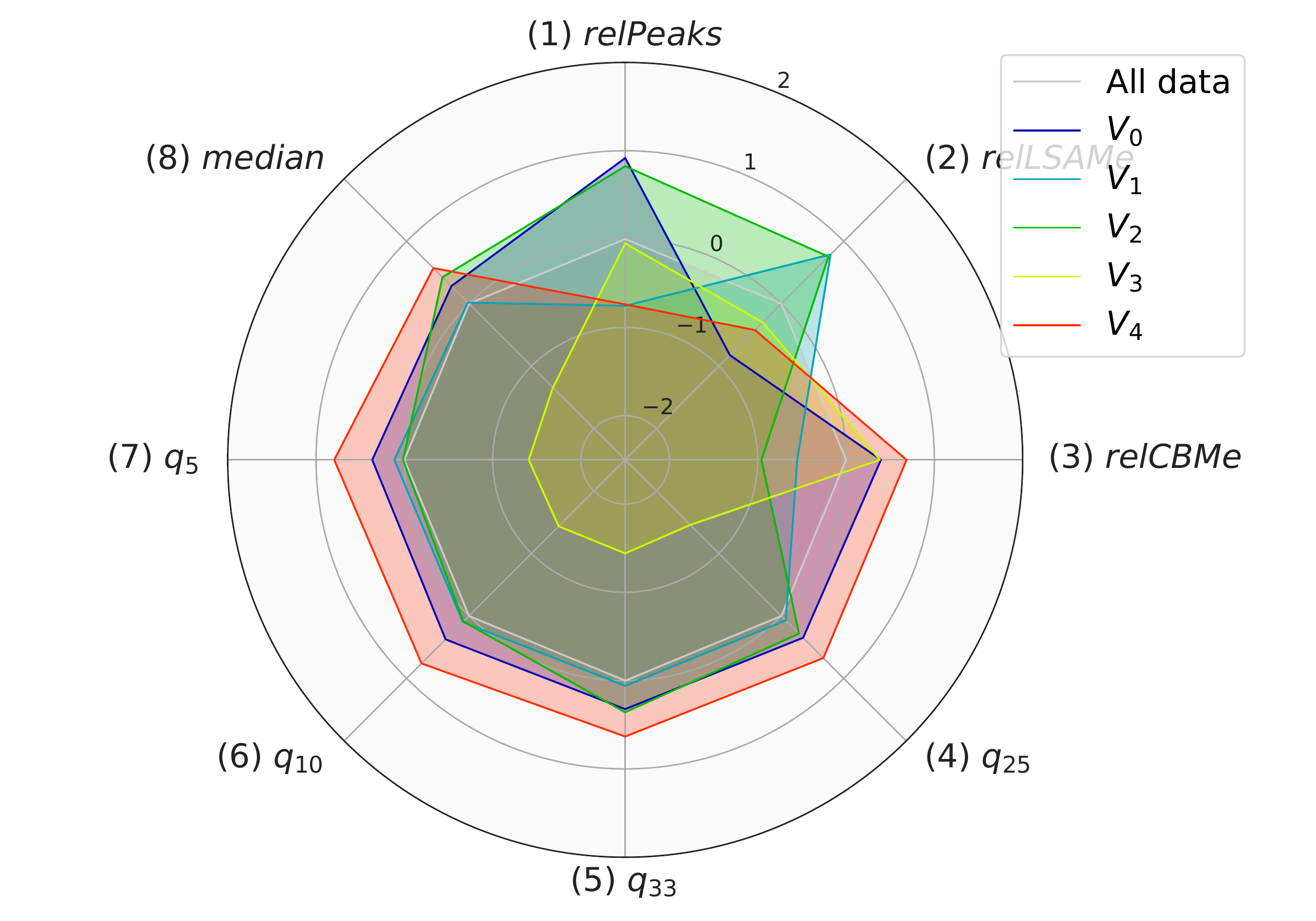}
        \caption[]%
        {{\small Distinctive features across all classes.}}    
        \label{fig:sub_radar}
    \end{subfigure}
    \hfill    
    \begin{subfigure}[b]{0.23\textwidth}   
        \centering 
        \includegraphics[trim={0 0 0 0},clip,width=\textwidth]{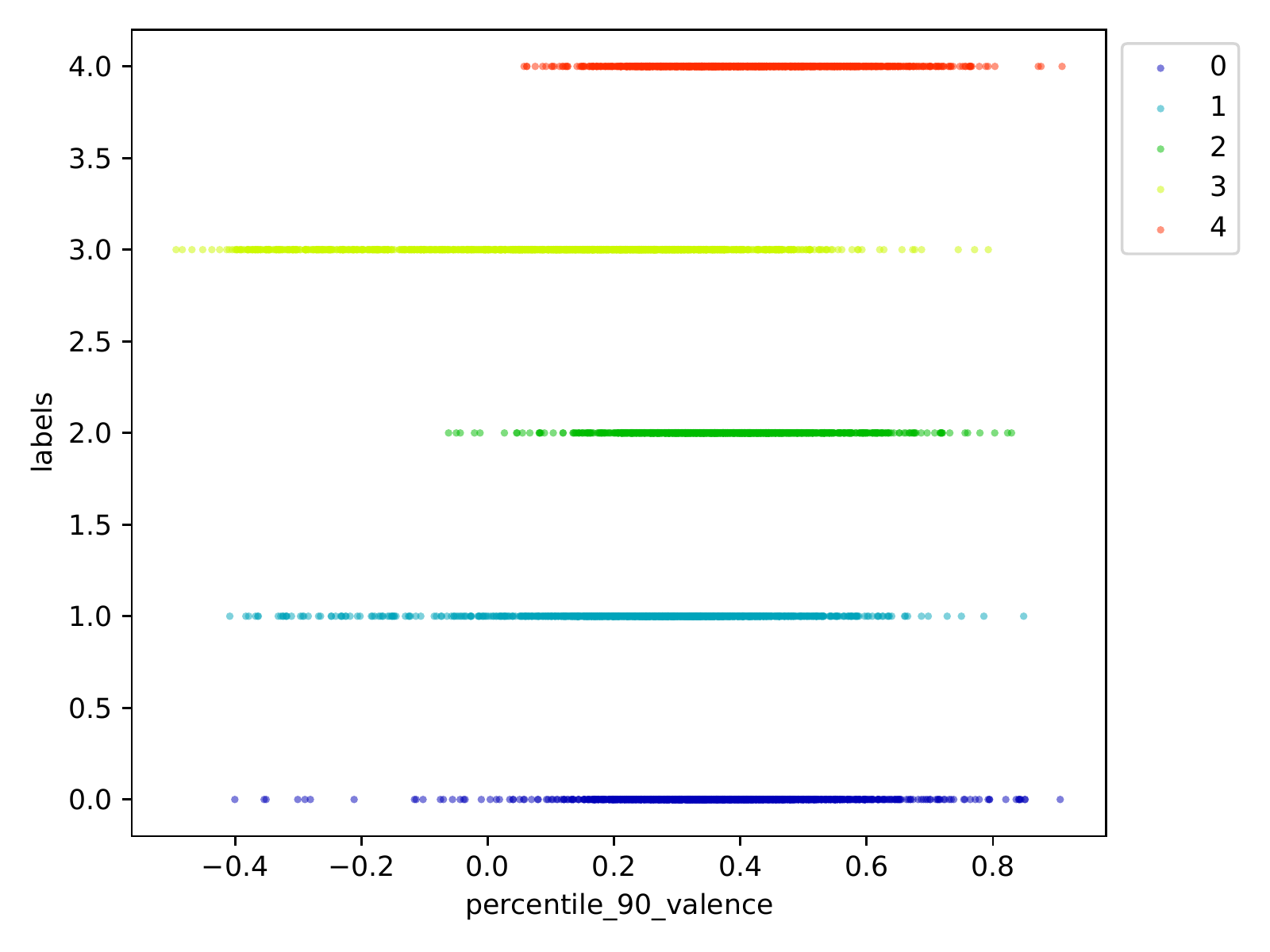}
        \caption[]%
        {{\small Example correlation between the labels and the 90th percentile.}}    
        \label{fig:sub_perentile}
    \end{subfigure}
    \caption[ The average and standard deviation of critical parameters ]
    {\small Exemplary visualisation capabilities of \name for the class extraction process.} 
    \label{fig:visual}
\end{figure*}

\vspace{-0.1cm}
\subsection{Emotional Class Extraction}
Clustering is by nature an unsupervised machine learning process, and so,  human monitoring of the found class clusters ensures they are based on meaningful patterns. The \name provides a number of tools for this purpose. After each clustering outcome, detailed profiling is carried out, which contains statistics, \eg mean and standard deviation, as well as visualisations of the obtained clusters. \Cref{fig:visual} summarises these: a) shows a correlation between each feature and the cluster classes. This aids identification of influential features. b) offers a visual interpretation of the clustered features through dimension reduction. c) provides an overview of the degrees of influence for individual features in the entire cluster class, ordered by the overall importance (distance from the average value across all classes), while d) shows the statistical (normalised) distribution of a single feature per class.

The outcome of clustering is highly dependent on the dataset, and specifically the distribution of underlying emotional annotations. For this reason, it is difficult to generalise the current findings. In the following, we summarise a few general tendencies that we observe from current experiments.

For this, we run experiments applying \kmeans, \cmeans, GMM, and \agglo
on \musesec. For the input features, we select four different feature sets: 
distribution-based features $set_{basic}$\footnote{mean, median, std., $q_{\{5, 10, 25, 33, 66, 75, 90, 95\}}$}, 
time-series features as in 
$set_{change}$\footnote{std., rel.\ energy, rel.\ sum of changes, rel.\ number peaks, rel.\ long strike below mean, rel.\ long strike above mean, rel.\ count below mean}, 
$set_{ext.}$\footnote{$set_{basic} \cup set_{change}$ + rel.\ number crossing 0, percentage of reoccurring data points to all data points}, 
and a very large feature set $set_{large}$\footnote{$set_{ext.}$ + skewness, kurtosis, mean abs.\ change, mean change, mean second derivative central, and the first and last location of the minimum and maximum, respectively}. We further explore the reducing the dimensions before the clustering setting the PCA parameter to \{None, 2, 5\}, and specify the number of clusters to \{3, 5\}.

We defined one criterion of a fruitful outcome, \ie if the cluster measures achieve optimal results (\cf \Cref{sec:measures} for difficulties). 
Furthermore, the identification of distinct cluster characteristics and a similar size of the classes may express optimal clusters. The experiments show that the composition of the features has a major influence on achieving the desired results. The features describing the distribution ($set_{basic}$) achieve slightly better results in terms of clustering measures than the feature set describing changes over time $set_{change}$. However, the latter seems to capture specific clusters very well, which is expressed by a small set of features (\cf \Cref{fig:sub_radar}) that stands out strongly from the average characteristics of these across all clusters. Mixing these two feature sets to the $set_{ext.}$ leads to the most evenly distributed class sizes. We recommend experimenting with the two general feature types and compiling your own set of reliable features for a given dataset, depending on your criteria and results obtained.

Regarding the class distribution, in 9 out of 96 setups created, at least one cluster does not cover enough percentage of the total amount of data points to fulfil our class-size-by-chance threshold of 25\,\%. With an increasing number of clusters (above five), all algorithms tend to split up existing smaller class clusters into even smaller ones, making it more likely to violate the class size rule. This behaviour occurs regardless of the feature set used.

In our feature reduction experiments, brute force was used to determine the best number of components. It showed that almost all clustering metrics except S\_dbw perform better when a two-component PCA is used before clustering. However, in terms of the ability to predict the generated class clusters, in our case, five components is the better choice (by chance level vs maximum result). Another decisive aspect in this process is the size (and types) of the feature sets to use for dimension reduction.
Prediction results obtained by using this process can be found in \cite{stappen2021muse}. 

Finally, we find two other high impact aspects noteworthy: the segment length and the data basis for clustering.
Regarding the segment length, the time series features (\eg long strike below mean) are sensitive to the length of the segment compared to the features that only describe the distribution (\eg quantile). If segments of varying length are given, it is recommended to adjust the length of the segments if possible and to convert the features by length from an absolute to a relative value corresponding to the length of the segment, avoiding the creation of meaningless classes.
For the affected features implemented in this toolkit, the normalisation by length is already performed by default.

Depending on the partitioning of the dataset, \ie the homogeneity between training, development, and test partitions, a clustering algorithm can generate completely different cluster classes. 
If the tool is used in the sense of an end-to-end process, where first the continuous signals are predicted and then a transformation into classes is automatically performed by a pre-trained clustering model, the exclusive use of the training dataset is advisable to test the method under real conditions. If it is a one-off process where suitable discrete classes are to be found for a given continuous annotation, the extraction can also be carried out on all data. 

Of further note, we have found that using DBSCAN\footnote{DBSCAN parameters: $\epsilon=\{0.01, 0.05, 0.1, 0.25\}$; min\_samples={3, 5}; PCA=\{None, 2, 5\}} for this task is less optimal. First, the class size threshold must be disabled because at least one resulting class does not meet the minimum size (\eg the noise cluster). Second, the algorithm tends to produce a very low (1-2) or very high number of classes (up to 300). 

\vspace{-0.1cm}
\section{Conclusions}
In this paper, we introduced the \name -- a novel annotation toolkit for creating continuous and discrete gold standards. It provides capabilities to compare different fusion strategies of continuous annotations to a gold standard as well as simplify this gold standard to classes by extracting and clustering temporary and local signal characteristics. Hence, we provided a unified way to create regression and classification targets for emotion recognition. Furthermore, we introduced \awe combining the annotation alignment on every time step and intelligently weighting of the individual annotation. Finally, important configuration parameter were highlighted in our series of experiments to which illustrated the toolkit's capabilities on the \musesec dataset. In the future, we plan to add further functionality, such as extending the dimension reduction to T-SNE and LDA.



\footnotesize
\bibliographystyle{ACM-Reference-Format}
\balance
\bibliography{sample-base}


\end{document}